\documentclass{article}

\usepackage{PRIMEarxiv}

\usepackage[utf8]{inputenc} 
\usepackage[T1]{fontenc}    
\usepackage{float}
\usepackage{hyperref}       
\usepackage{url}            
\usepackage{booktabs}       
\usepackage{amsfonts}       
\usepackage{nicefrac}       
\usepackage{microtype}      
\usepackage{lipsum}
\usepackage{fancyhdr}       
\usepackage{graphicx}       
\usepackage{amsmath}
\graphicspath{{media/}}     
\usepackage[table]{xcolor}
\usepackage{adjustbox}  
\usepackage{natbib}

\usepackage[scaled]{helvet} 
 
\usepackage{mathpazo} 
\title{VisioFirm: Cross-Platform AI-assisted Annotation Tool for Computer Vision}
\author{
  Safouane EL GHAZOUALI*\\
  TOELT LLC AI lab\\
  Winterthur, Swintzerland \\
  \texttt{safouane.elghazouali@toelt.ai} \\
  \And
  Umberto Michelucci \\
  TOELT LLC AI lab, \\
  Winterthur, Swintzerland \\
  \texttt{umberto.michelucci@toelt.ai} \\
}

\begin{document}
\maketitle


\begin{abstract} 

    AI models rely on annotated data to learn pattern and perform prediction. Annotation is usually a labor-intensive step that require associating labels ranging from a simple classification label to more complex tasks such as object detection, oriented bounding box estimation, and instance segmentation. Traditional tools often require extensive manual input, limiting scalability for large datasets. To address this, we introduce VisioFirm, an open-source web application designed to streamline image labeling through AI-assisted automation. VisioFirm integrates state-of-the-art foundation models into an interface with a filtering pipeline to reduce human-in-the-loop efforts. This hybrid approach employs CLIP combined with pre-trained detectors like Ultralytics models for common classes and zero-shot models such as Grounding DINO for custom labels, generating initial annotations with low-confidence thresholding to maximize recall. Through this framework, when tested on COCO-type of classes, initial prediction have been proven to be mostly correct though the users can refine these via interactive tools supporting bounding boxes, oriented bounding boxes, and polygons. Additionally, VisioFirm has on-the-fly segmentation powered by Segment Anything accelerated through WebGPU for browser-side efficiency. The tool supports multiple export formats (YOLO, COCO, Pascal VOC, CSV) and operates offline after model caching, enhancing accessibility. VisioFirm demonstrates up to 90\% reduction in manual effort through benchmarks on diverse datasets, while maintaining high annotation accuracy via clustering of connected CLIP-based disambiguate components and IoU-graph for redundant detection suppression. VisioFirm can be accessed from \href{https://github.com/OschAI/VisioFirm}{https://github.com/OschAI/VisioFirm}.
    
\end{abstract}

\section{Introduction}\label{sec:intro}

In the field of computer vision (CV), high quality datasets are essential for training robust machine learning (ML) models in tasks such as object detection and instance segmentation. These annotations provide the necessary ground truth for supervised learning algorithms to learn patterns and generalize accross other similar images. However, the process of data annotation remains one of the most hard stages in CV pipelines, often consuming a significant portion of project resources including time. With the  growth of image data from divers sources like autonomous vehicles \cite{liu2024surveyautonomousdrivingdatasets}, medical imaging \cite{Jim_nez_S_nchez_2025}, surveillance systems\cite{An2025surveillance} and remote sensing \cite{velazquez2025earthview}, the demand for efficient annotation tools has also grown.

Data annotation presents several challenges that obstruct scalability and efficiency. When performed manually, this process can be time consuming and labor intensive due to the requirement of human annotators to accurately label objects, boundaries, and attributes in massive amounts of images. Common issues include annotation errors due to subjective interpretations, data biases introduced by inconsistent labeling practices, and difficulties in scaling to large datasets without incurring prohibitive costs \cite{ahmadzadeh2025guidemanualannotationscientific}. Additionally, when handling sensitive data, such as in medical or surveillance applications \cite{Yadav2023-zx} security and privacy concerns arise. Moreover, traditional manual annotation methods may struggle with complex tasks like objects with complex shapes and textures, tasks requiring oriented bounding boxes such small objects in remote sensing \cite{Wang_2025}, where precision is important. Manual performances can lead to inaccuracies or a lot of subjectivity which might corrupt to a degree the dataset quality.
Some existing annotation tools attempt to mitigate these issues but often fall short in some aspects. Softwares such as VIA (VGG Image Annotator) \cite{Dutta_2019} provide basic manual labeling capabilities for rectangles and polygons with the possibility of tracking some bounding box in subsequent frames of a video but lack several automations, making them unsuitable for large-scale projects. Several more advanced platforms such as LabelStudio \cite{Labelstudiosoft}, CVAT \cite{boris_sekachev_2020_4009388}, and Roboflow \cite{ciaglia2022roboflow100richmultidomain}, can incorporate AI assistance to speed up the process. However, some may require either cloud dependencies, subscription fees, additionally dependencies or specialized hardware.

To address these gaps, we present VisioFirm, an open-source, cross-platform web application that integrates advanced AI models to automate to the best image labeling for CV tasks. We aim at Object detection, oriented object detection and segmentation subfields. VisioFirm uses a filtering system combined with low thresholding inferencers from pretrained and zeroshot model. The hybrid system combines CLIP \cite{radford2021learningtransferablevisualmodels} for semantic filtering with detectors for standard classes and zero-shot capabilities via Grounding DINO for custom ones. This approach generates high-recall pre-annotations that can be auto-refines through several post-processing step, but the user can further refine through an intuitive interface supporting bounding boxes, oriented bounding boxes, polygons, and on-the-fly segmentation using Segment Anything Model (SAM) \cite{ravi2024sam2segmentimages} accelerated by WebGPU. The tool operates offline after initial model downloads, supports multiple export formats, and has been benchmarked to reduce manual effort by up to 90\% on diverse datasets while preserving annotation quality.

The contributions of this work are threefold:
\begin{itemize}
    \item[(1)] A user friendly web-application, platform agnostic and easy to install;
    
    \item[(2)] A novel AI-assisted annotation pipeline that enhances efficiency through low-threshold pre-annotations and CLIP-based filtering;
    
    \item[(3)] An accessible, browser-based tool with specialized support for oriented object detection and segmentation;
\end{itemize}



\section{VisioFirm}\label{sec:visiofirm}

VisioFirm is built on a Flask Python for backend logic and uses a HTML-CSS-JavaScript frontend for interactive user interfaces. The tool supports user authentication, project management, and easiness in handling large batches of image datasets. Key features include keyboard shortcuts for rapid navigation, zoom/pan controls, and multiple annotation modes (rectangle, polygon, magic wand for segmentation). Annotations can be exported in standard formats such as YOLO, COCO, Pascal VOC, and CSV, to enable integration with other CV frameworks.

\begin{figure}[htp!]
    \centering
    \includegraphics[width=\linewidth]{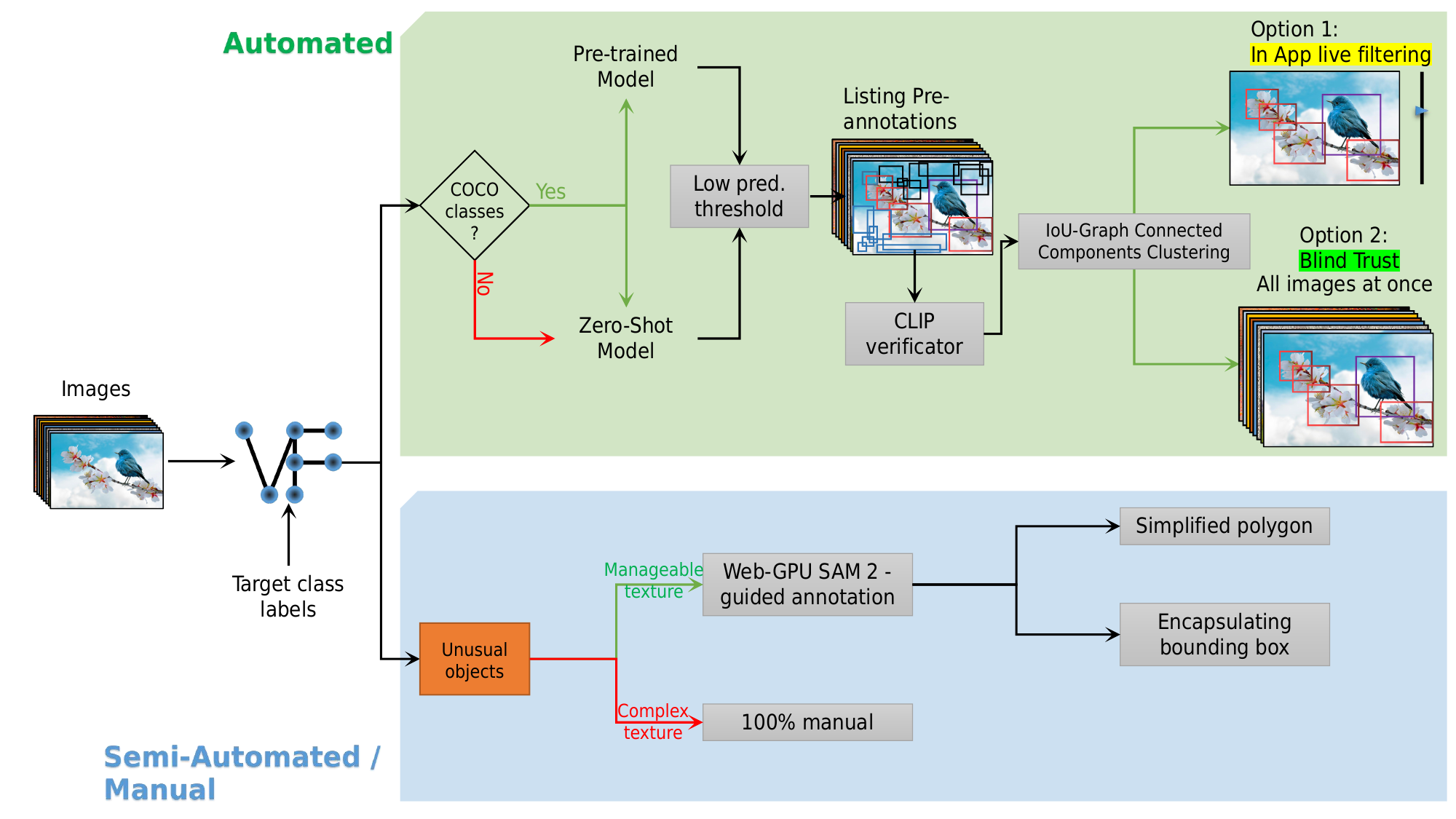}
    \caption{Illustration of the overall VisioFirm pipeline with both automated and semi-automated or fully manual annotation.}
    \label{fig:visiofirm_pipeline}
\end{figure}
\newpage

The overall workflow of VisioFirm is illustrated in Figure \ref{fig:visiofirm_pipeline}, which shows the hybrid labelling approach combining automated pre-annotation with semi-automated and manual refinement options. Starting from input images and target class labels, the system branches based on user usecase:

\begin{itemize}

    \item For classes aligned with common datasets such as COCO, VisioFirm leverages pretrained detection models to generate initial bounding box proposals. A low prediction confidence threshold (e.g., 0.2) is applied to prioritize recall, ensuring a broad capture of potential objects for subsequent refinement. The Ultralytics framework \cite{ultralytics2023yolo} is selected for its integration easiness in the ML-powered backend. Additionally, it offers robust support for multiple models as well as custom training and deployment. By default, YOLOv10 \cite{wang2024yolov10realtimeendtoendobject} (the largest model) is employed due to its state-of-the-art performance, eliminating the need for Non-Maximum Suppression (NMS) and achieving a favorable accuracy-latency trade-off. However, YOLO10 can be substituted with other Ultralytics-compatible architectures (RT-DETR, YOLOv5, YOLOv11 ...etc) or a custom-trained model to suit specific dataset characteristics.
    
    \item For custom or domain-specific classes not covered by pretrained models, VisioFirm employs a zero-shot detection approach using Grounding DINO \cite{liu2024groundingdinomarryingdino}. Grounding DINO is transformer-based model, enhanced by grounded pre-training, enables open-vocabulary detection with a reported 52.5 AP on COCO zero-shot transfer, scalable to 63.0 AP with fine-tuning. It is able to process text prompts to identify objects without requiring retraining, making it ideal for unusual or user-defined categories. The model integrates with VisioFirm's pipeline to generate initial annotations, which are then verified for accuracy, supporting flexibility in handling diverse annotation tasks.
    
    \item Pretrained and zero-shot models may underperform when encountering domain-specific or unusual objects, particularly those with complex shapes, textures, or occlusions not represented in training datasets like COCO. Recent studies highlight limitations in generalization, where models like YOLOv10 or Grounding DINO as a zero-shot model may produce low-confidence or erroneous detections for niche objects. VisioFirm addresses this by dynamically assessing detection quality, flagging cases where automated methods fall short, thus triggering a transition to AI-assisted annotation. For this, we use a web-gpu segment anything 2 \cite{ravi2024sam2segmentimages} model for single-click annotation.
    
    \item In case the previous 3 options were not successful at retrieving label automatically, VisioFirm provides a manual annotation fallback that the use can select. Inspired by existing tool such as CVAT, VAT and LabelStudio, all the basic functionalities for manual labelling have been designed in a user-friendly interface.
    
\end{itemize}

These initial detections undergo filtering via IoU-Graph Connected Components Clustering and CLIP verification to remove redundancies and ensure label accuracy. Users can then opt for in-app live filtering (Option 1) or blind trust of all pre-annotations (Option 2). For complex scenarios, such as unusual objects or textures, the pipeline shifts to semi-automated modes using WebGPU-accelerated SAM2 for guided segmentation, or fully manual annotation as a fallback. Post-processing includes contour simplification for polygons and encapsulation into bounding boxes where needed.

\subsection{Image annotation}

The image annotation module in VisioFirm provides a robust, interactive canvas for manual and semi-automated labeling, built on HTML5 Canvas with JavaScript handlers for drawing, interaction, and state management. Users load images via a dashboard, where they can navigate through datasets as illustrated in Figure \ref{fig:visiofirm_web_interface}. Annotation tools include:

\begin{figure}[htp]
    \centering
    \includegraphics[width=\linewidth]{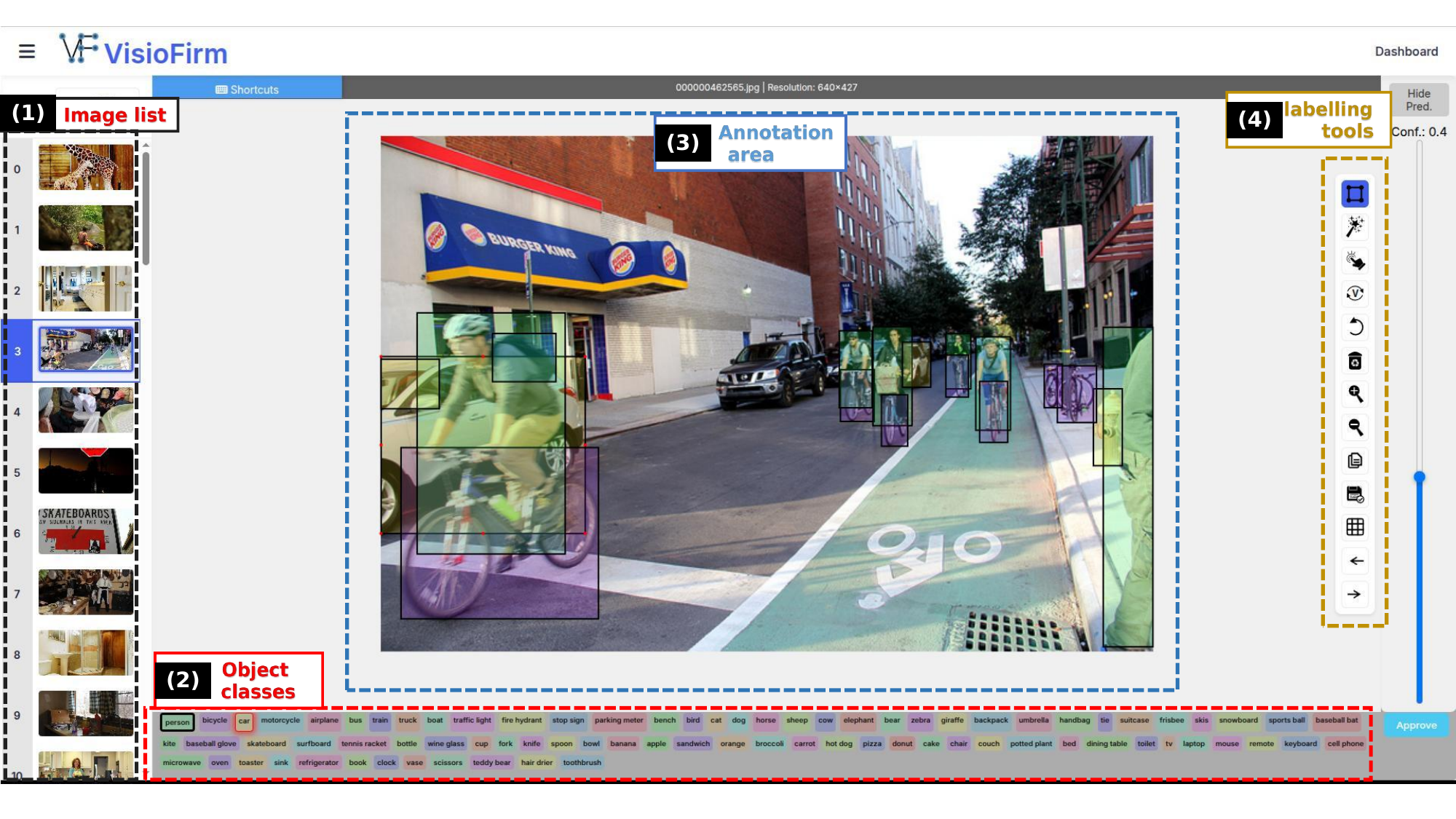}
    \caption{Visualization of the VisioFirm interface of the annotation. The figure shows bounding box project and 4 interface branches: (1) the image list is visualized at the left side for navigation. (2) the list of object classes defined at the project creation. (3) Annotation area featuring the image with zoom-in and out possibility. (4) labelling helper buttons with custom options like activate of magic mode, reset, undo, set default view, delete annotation, duplication and save.}
    \label{fig:visiofirm_web_interface}
\end{figure}

Depending on the selected project setup, VisioFirm offer an access to 3 labelling modes: (1) \textbf{Rectangle Mode}: For standard bounding boxes, supporting drag-and-resize with real-time coordinates. (2) \textbf{Polygon Mode}: For precise segmentation boundaries, allowing point-by-point drawing with auto-closure and editing. (3) \textbf{Oriented Bounding Box (OBB) Mode}: Enabling rotated rectangles for aligned object representation, computed via minimum area encapsulation.
All these can also be automated via the fourth option (4) \textbf{Magic Mode}, which integrates SAM2 for click-guided segmentation, accelerated by WebGPU to ensure low-latency performance in the browser. The annotation interface offers also shortcut for fast and efficient annotation.

Annotations are stored locally in a SQLite database for persistence and can be refined with features like duplication, deletion, and confidence adjustments. For semi-automated workflows, pre-annotations from the VFPreAnnotator are overlaid, allowing users to accept, modify, or reject proposals with minimal input.

\subsection{VisioFirm Pre-annotator}
\begin{figure}
    \centering
    \includegraphics[width=\linewidth]{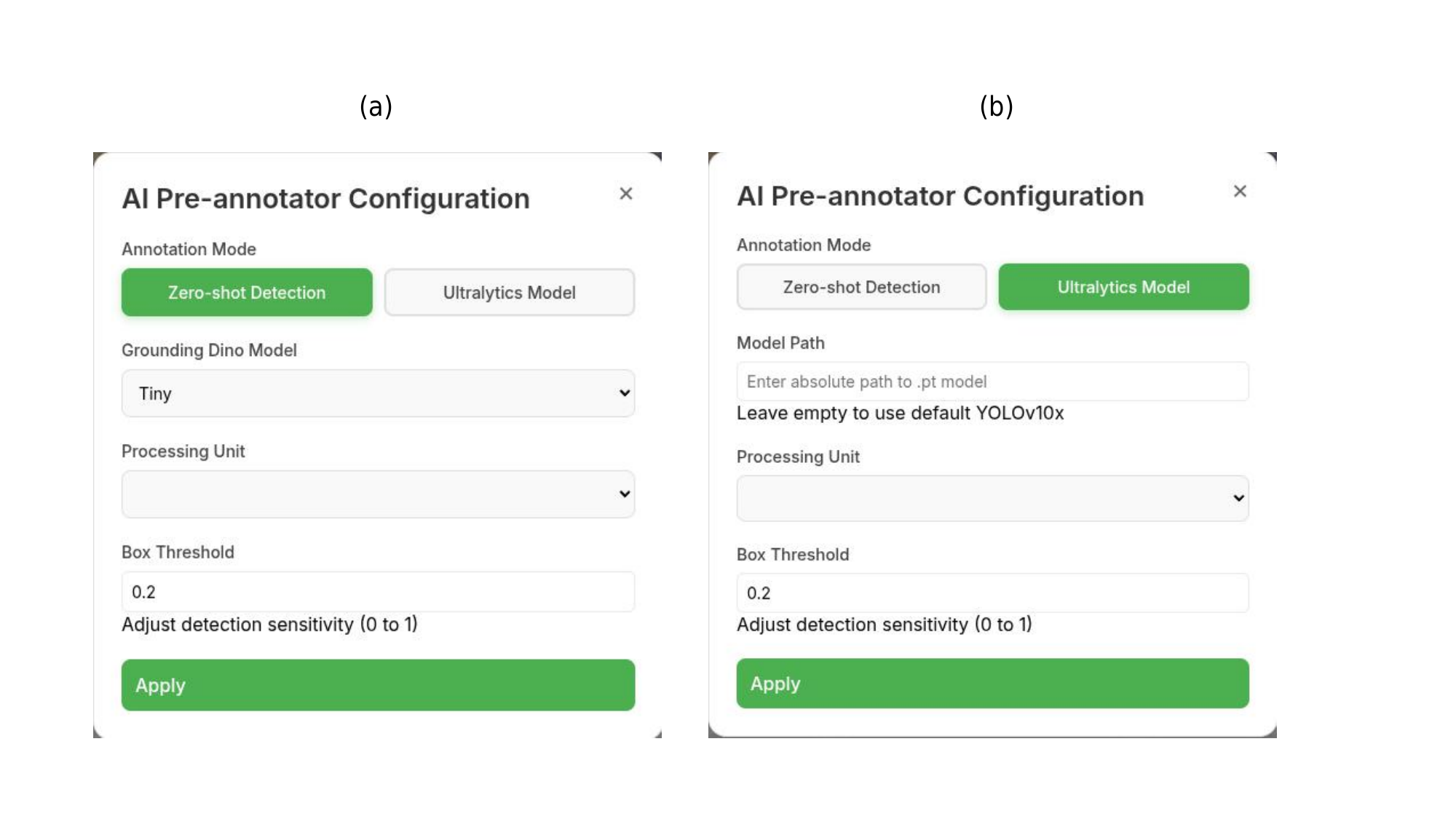}
    \caption{VisioFirm Pre-annotator configuration: (a) using zero-shot Grounding Dino model. (b) using Ultralytics-compatible models defaulted to YOLOv10x}
    \label{fig:preannotator_config}
\end{figure}

The VFPreAnnotator is the core automation component of VisioFirm, designed to generate high-recall initial annotations using a hybrid detection pipeline. Implemented in Python with libraries like Ultralytics, Transformers, and CLIP, it processes images in batch mode, querying a project database for classes and paths. Figure  \ref{fig:preannotator_config} is a visual of the Pre-annotator options offer in VisioFirm app.

The pipeline begins by parsing target classes and selecting the appropriate model: pretrained YOLOv10 \cite{wang2024yolov10realtimeendtoendobject} for COCO-aligned classes or zero-shot Grounding DINO for custom ones, both with a low confidence threshold (such as 10\% or less) to maximize recall. Detections are batched for efficiency, with post-processing to clean labels and filter invalid outputs. To verify and refine labels, especially in cases of potential mismatches from low-threshold predictions, the pipeline crops the image region defined by each predicted bounding box \( b_k = [x_{k1}, y_{k1}, x_{k2}, y_{k2}] \), extracting a sub-image \( I_k \). This crop is then fed into CLIP for semantic verification against the set of target classes \( \mathcal{L} \).

CLIP computes normalized embeddings: \( \mathbf{f}(I_k) \) for the image crop and \( \mathbf{g}(l) \) for each class label \( l \in \mathcal{L} \). Cosine similarities are calculated using Eq. \ref{eq:s_j}. Afterward, a softmax is applied to convert the values into probabilities as in Eq. \ref{eq:clip_softmax} 

\begin{equation}
    s_j = \mathbf{f}(I_k) \cdot \mathbf{g}(l_j),
\end{equation}\label{eq:s_j}

\begin{equation}
    p_j = \frac{\exp(s_j / \tau)}{\sum_{m=1}^{|\mathcal{L}|} \exp(s_m / \tau)},
\end{equation}\label{eq:clip_softmax}
where \( \tau \) is a temperature parameter (default 1 in the implementation). 

Finally: The best label is selected as:
\begin{equation}
    l^* = \arg\max_{j} p_j.
    \end{equation}\label{eq:clip_argmax}

This CLIP-based verification guarantees the semantic alignment between the predicted label and the cropped content from the original image. It then rectifies false positives from the initial detector while exploring the low-threshold approach for broad coverage. At the end of the pre-annotation stage, candidate annotations will be summarized and temporarily stored for further filtering and user refinement in the app (e.g., minimum-confidence setting, manual override). In segmentation mode, our SAM2 policy casts the validated bounding boxes as the initialization masks to be refined in the post-processing. This balanced model gives preference to recall in generation and precision in verification, which has been shown to save manual labor in the experiments.

\subsection{Pipeline Filtering and Post-Processing}
The VisioFirm pipeline employs filtering and post-processing techniques to refine the pre-annotations, removing redundancies and enhancing quality. Redundancies among the listed pre-annotations are addressed through IoU-Graph Connected Components Clustering: A graph \( G \) is constructed where nodes represent detections, and edges connect pairs with an Intersection over Union (IoU) greater than 0.9, calculated as:

\begin{equation}
    \text{IoU}(b_i, b_j) = \frac{ | b_i \cap b_j | }{ | b_i \cup b_j | },
\end{equation}\label{eq:iou_filter}

where \( b_i \) and \( b_j \) are bounding box coordinates, with intersection and union derived from min-max overlaps.

Connected components are identified using a graph traversal algorithm (e.g., DFS or BFS), grouping overlapping detections into clusters. Within each cluster, if labels are consistent, the detection with the highest confidence score is retained. For clusters with conflicting labels (post-CLIP verification), a union bounding box is computed as \( b_u = [\min x_1, \min y_1, \max x_2, \max y_2] \), and a final CLIP check on the union crop \( I_u \) resolves the best label, retaining the top-scoring matching detection.

\begin{figure}[htp!]
    \centering
    \includegraphics[width=\linewidth]{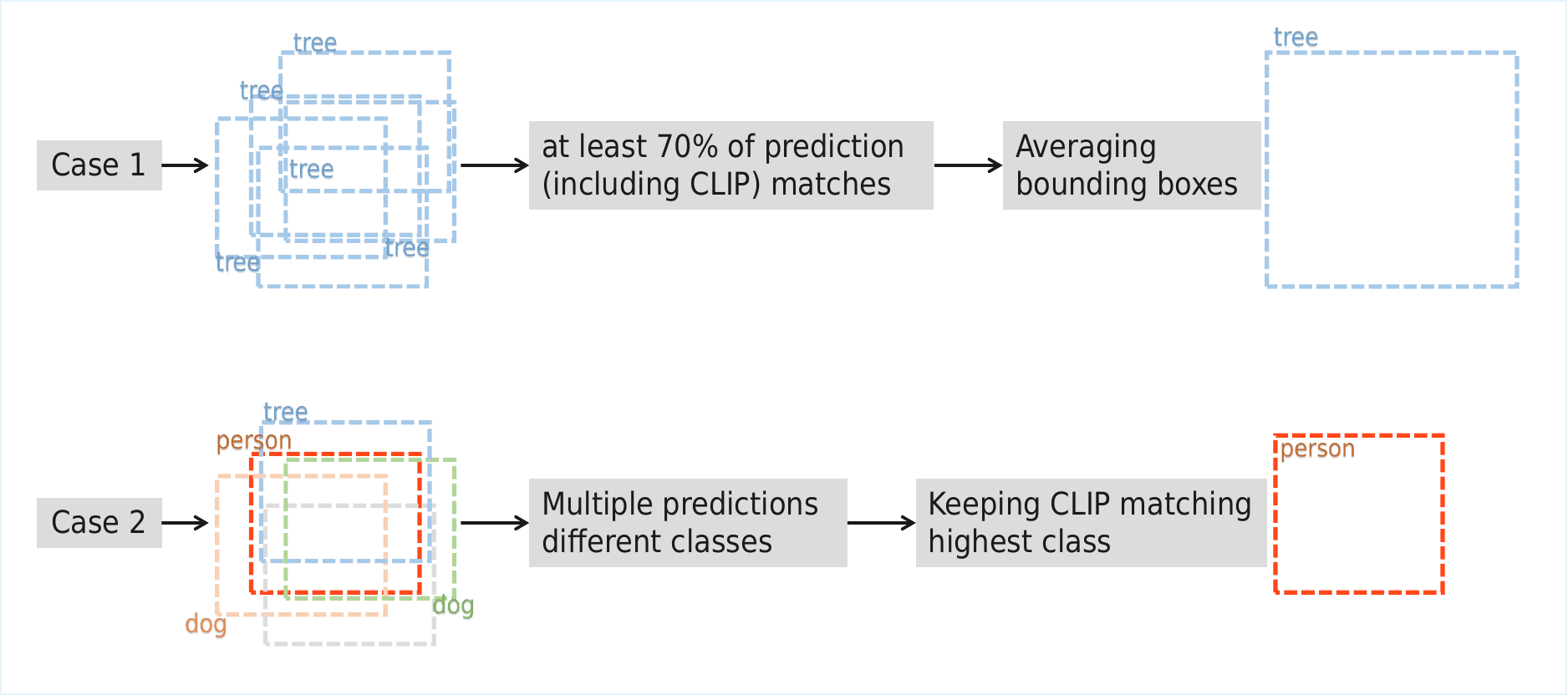}
    \caption{VisioFirm post-processing applied as an output of Pre-annotation pipeline for fast and automated detection. Case 1: the predictions are almost grouping toward the same label with some differences in the labels sizes. Case 2: Multiple different predictions with IoU over 70\% means that an object of interest is whithin that area and thus reliance on post processing to assign a label.}
    \label{fig:casespreannotation}
\end{figure}

An example illustration of the filtering and post-processing pipeline cases can be viewed in Figure \ref{fig:casespreannotation}.

For segmentation tasks, post-processing includes adaptive hole-filling via morphological closure (iterative closing with increasing kernel size up to 10 iterations) and contour simplification using the Ramer-Douglas-Peucker algorithm \cite{douglas1973algorithms} $\epsilon = 0.002 \times \text{perimeter}$ reducing polygon points while preserving shape (minimum 3 points required). Valid annotations are inserted into an SQL database, including image metadata, label identifiers, and geometry data (rectangles or polygons).

The pipeline dynamically assesses reliability; if automated pre-annotation underperforms on unusual objects or complex textures, it triggers WebGPU-accelerated SAM2 for guided annotation or switches to 100\% manual mode, as shown in Figure \ref{fig:visiofirm_pipeline}. This ensures high annotation accuracy with minimal redundancy.

\subsection{Graph Construction}

To model relationships between the forbidden content chunks, a $k$-nearest neighbors (kNN) graph $\mathcal{G} = (\mathcal{V}, \mathcal{E})$ is constructed over the embeddings. The vertices $\mathcal{V} = \{1, \dots, N\}$ correspond to the individual chunks. Edges are established between each node and its $k_g$ closest neighbors, where $k_g$ is a hyperparameter determining the graph's connectivity (e.g., $k_g=50$ to ensure sufficient local structure without excessive density). Self-connections are excluded to avoid trivial loops.

The similarity between two embeddings $\mathbf{e}_i$ and $\mathbf{e}_j$ is computed using cosine similarity, and the edge weight $w_{ij}$ is defined as $w_{ij} = 1 - \cos^{-1}(\mathbf{e}_i^\top \mathbf{e}_j)/\pi$, which converts the cosine distance to a similarity score in [0,1]. The adjacency matrix $\mathbf{A} \in \mathbb{R}^{N \times N}$ of the graph is then expressed by Eq. \ref{eq:adj-matrix}:

\begin{equation}
    A_{ij} = 
    \begin{cases} 
    w_{ij} & \text{if } j \in \mathcal{N}_i(k_g), \\
    0 & \text{otherwise},
    \end{cases}
    \label{eq:adj-matrix}
\end{equation}

where $\mathcal{N}_i(k_g)$ denotes the set of the $k_g$ nearest neighbors of node $i$ based on the similarity metric.

This graph represents semantic connections within the forbidden topics, enabling the propagation of relevance signals through diffusion processes like PageRank. A sparser graph (controlled by $k_g$) reduces computational overhead while capturing essential clusters of related content.

\section{Web Application Design}\label{sec:webdesign}

VisioFirm's web application architecture has been designed such to prioritize computational flexibility, performance optimization, and user-centric analytics in computer vision annotation tasks. The system enables hardware-agnostic deployment, allowing users to specify exclusive CPU execution for resource-constrained environments or leverage GPU acceleration (via CUDA for backend inference or WebGPU for client-side segmentation) when available. This configurability is implemented through runtime flags in the Flask backend, ensuring adaptability to diverse hardware profiles while maintaining consistent annotation accuracy.

To quantitatively assess the impact of hardware acceleration on pipeline efficiency, some experiments were ran on a benchmark dataset comprising 100 subset images from the COCO validation set, focusing on common classes for YOLOv10 inference. Experiments were for segmentation end task using an Intel Core i9-12th-12900k CPU and an NVIDIA RTX A6000 GPU (48 GB VRAM). Key metrics included total and per-image inference latency (including post-processing) at low (0\%) and higher (50\%) confidence thresholds, computed across the full pipeline.

\textbf{Inference Latency for YOLOv10 (Largest Variant)}: At 0\% threshold, CPU mode averaged 7.43 s per image (total: 12 min 23 s), while GPU mode reduced this to 2.53 s per image (total: 4 min 13 s), a 2.9x speedup. At 50\% threshold, CPU averaged 2.38 s per image (total: 3 min 58 s), with GPU at 0.14 s per image (total: 14 s), a 17x speedup.

\textbf{Inference Latency for Grounding DINO (Tiny Variant)}: At 0\% threshold, CPU averaged 29.86 s per image (total: 49 min 46 s), while GPU averaged 5.28 s per image (total: 8 min 48 s), a 5.7x speedup. At 50\% threshold, CPU averaged 9.03 s per image (total: 15 min 3 s), with GPU at 2.18 s per image (total: 3 min 38 s), a 4.1x speedup.

\begin{figure}[htp]
    \centering
    \includegraphics[width=\linewidth]{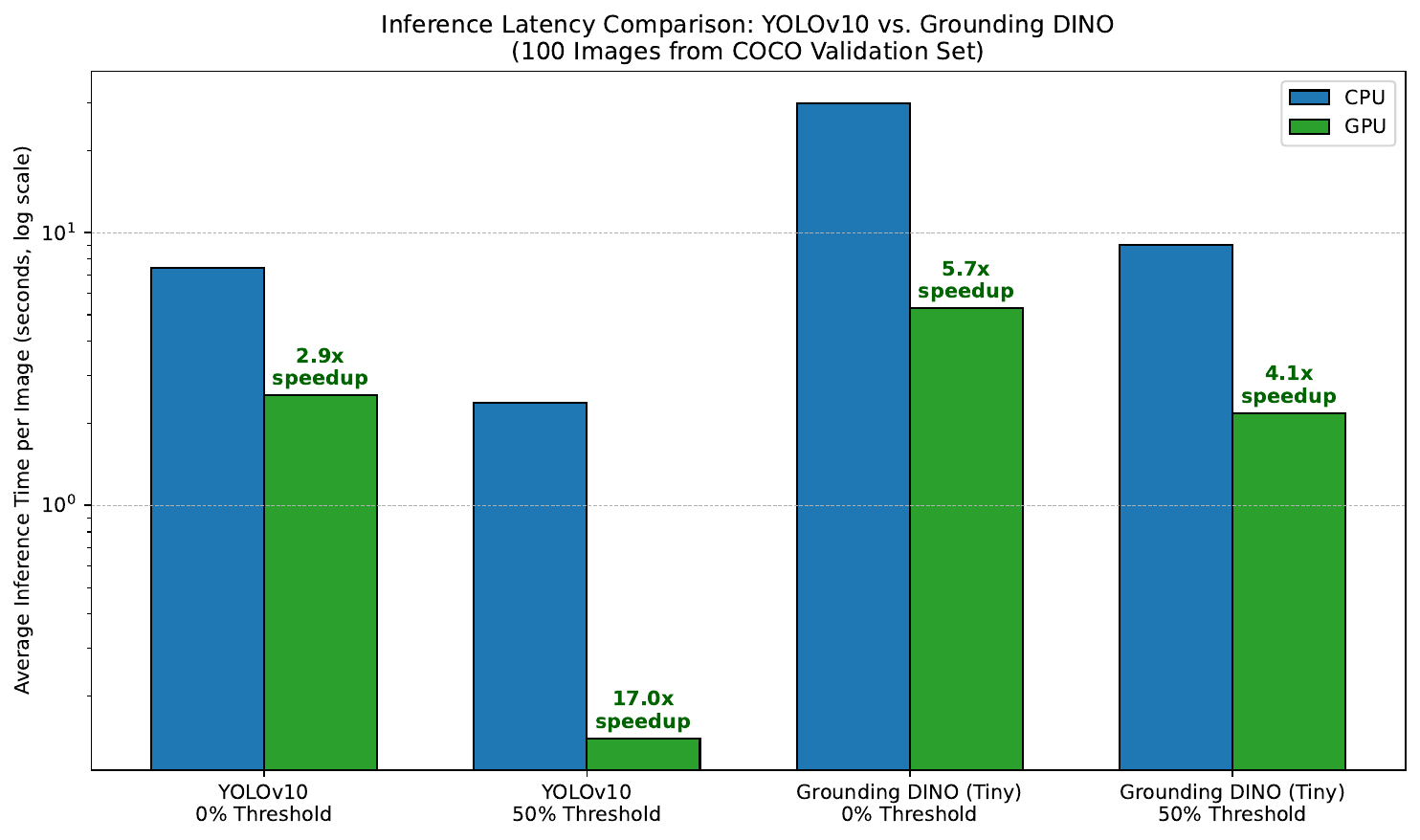}
    \caption{Bar chart comparing average per-image inference times (log scale) for YOLOv10 and Grounding DINO across thresholds and hardware, demonstrating GPU speedups and threshold impacts.}
    \label{fig:performance_comparison}
\end{figure}

These results, visualized in Figure \ref{fig:performance_comparison}, highlight significant speedups with GPU acceleration and higher thresholds, which reduce the number of detections processed in filtering stages. Though this omits many labels proposals, it can be very efficient for domain-specific models such as YOLO finetuned on specific image datasets for limited objects. Annotation precision (mAP@0.5) remained relatively consistent with some duplicate labels, especially when performing segmentation using GroundingDino Tiny combined with SAM2 for auto mask labels generation. For datasets with COCO-aligned classes, YOLOv10 is substantially faster than Grounding DINO (e.g., up to 15x at low thresholds on GPU), making it the preferred choice for common object annotation to optimize efficiency without compromising recall. On the other hand, Grounding DINO can be very efficient at generating uncommon class labels proposals even though it takes more time to compute on the full dataset.

VisioFirm supports interoperability as well, where user can import during the project creation, partially annotated datasets from external tools as long as the annotation conform to standardized formats COCO JSON \cite{lin2014microsoft} or YOLO TXT YAML. Imported annotations are parsed, processed and merged into the SQLite database proper to the project, and visualized for refinement, ensuring seamless integration with other annotation tools like LabelStudio \cite{Labelstudiosoft} or CVAT \cite{boris_sekachev_2020_4009388}.

\begin{figure}[htp]
    \centering
    \includegraphics[width=\linewidth]{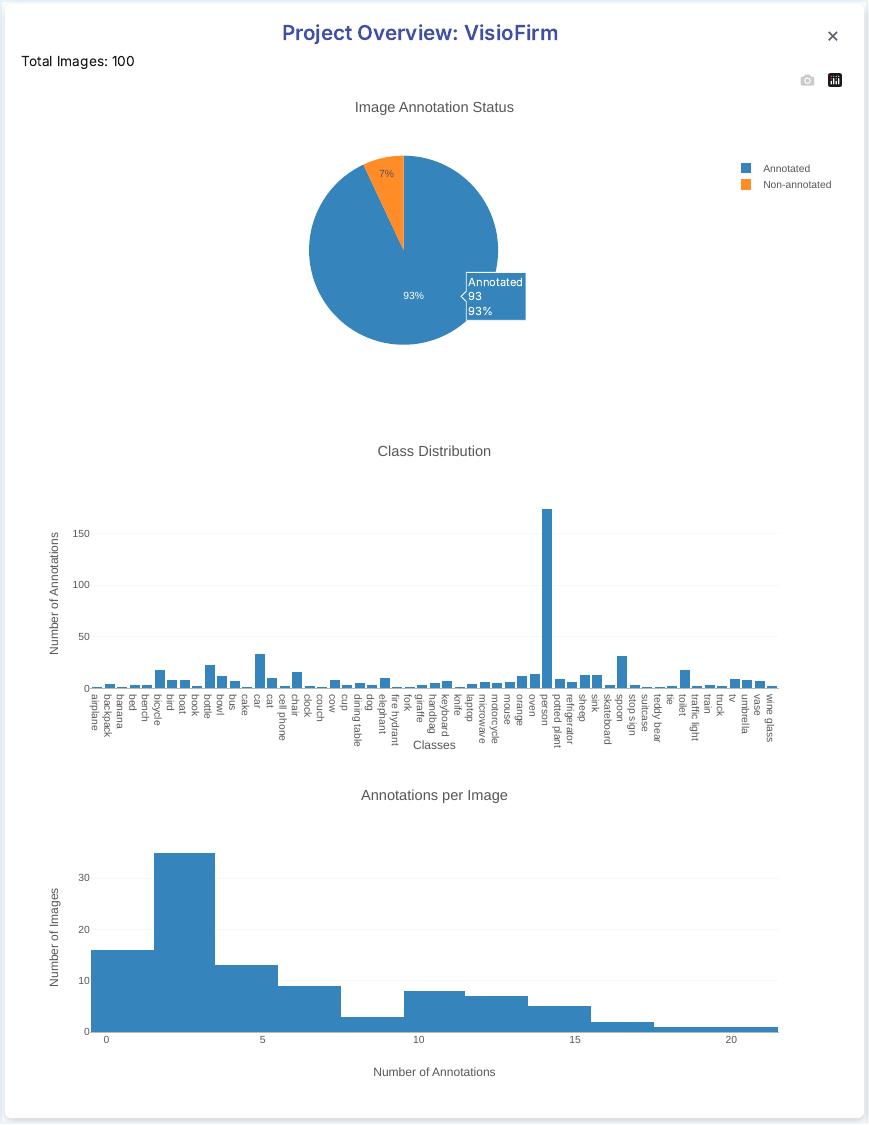}
    \caption{Example of VisioFirm project overview dashboard, displaying annotation status via pie chart, class distribution through bar chart, and annotations per image histogram.}
    \label{fig:project_overview}
\end{figure}

Furthermore, the application incorporates a dynamic progress monitoring module, generating analytical visualizations to inform annotation strategies. As illustrated in Figure \ref{fig:project_overview}, the dashboard presents:
- A pie chart depicting annotation completion status (e.g., 93\% annotated across 100 images).
- A bar chart of class distribution, revealing imbalances (e.g., high frequency for ``motorcycle'' versus sparse classes like ``vase'').
- A histogram of annotations per image, highlighting long-tail distributions (e.g., modal value at 0-5 annotations, with maxima exceeding 20).

These metrics, rendered via Plotly.js, enable data-driven decisions, such as prioritizing under-annotated classes or images, thereby mitigating biases in downstream model training.

This design framework enhances VisioFirm's utility in scientific and practical CV applications, balancing computational efficiency with analytical insights.

\section{Open Source Implementation and Dependencies}\label{sec:opensource}

VisioFirm is distributed as an open-source framework under the MIT License to foster collaborative advancements in computer vision annotation methodologies. The complete source code, including backend and frontend components, is hosted on GitHub at \url{https://github.com/OschAI/VisioFirm}, accompanied by comprehensive documentation, installation guides, and contribution protocols. This repository facilitates reproducibility, with versioned releases ensuring stable access for research replication.

The implementation leverages Python for backend processing, with compatibility with versions starting from 3.8. Table \ref{tab:dependencies} summarizes the core components, framework dependencies and their specifications.

\begin{table}[htp]
\centering
\small
\caption{Summary of VisioFirm's Software Dependencies and Versions}
\label{tab:dependencies}
\begin{tabular}{l l p{5cm}}
    \hline
    \textbf{Component} & \textbf{Version} & \textbf{Description} \\
    \hline
    Python & 3.8+ & Core app designed on Python 3.10. \\
    Flask & 2.0+ & Web framework for routing, authentication, and API endpoints. \\
    Ultralytics & 8.0+ & Library for YOLO model integration and object detection inference. \\
    Transformers & 4.30+ & Hugging Face library for Grounding DINO and zero-shot models. \\
    OpenAI-CLIP & Latest & Semantic embedding model for label verification and filtering. \\
    Torch & 2.0+ & PyTorch framework for deep learning operations (CPU/GPU support). \\
    NumPy & 1.24+ & Numerical computing for array manipulations in post-processing. \\
    OpenCV & 4.8+ & Computer vision library for image processing and contour handling. \\
    NetworkX & 3.0+ & Graph library for IoU-based clustering algorithms. \\
    SQLite3 & Standard & Embedded database for local annotation storage and queries. \\
    \hline
\end{tabular}
\end{table}

Installation can be executed in a fully automated way via \texttt{pip install -U visiofirm}, followed by \texttt{visiofirm} command line in a terminal to initiate the local server. For GPU-accelerated deployments, CUDA 12.0+ and compatible drivers are recommended; otherwise, the system defaultely falls back to CPU execution. Community contributions are solicited through pull requests, with emphasis on extending model compatibility (e.g., integration of emerging detectors) and enhancing cross-platform robustness, currently validated on Windows, macOS, and Linux systems.

\section{Conclusion and future direction}\label{sec:conclusion}

In this work, VisioFirm labelling tool is introduced, an open-source, cross-platform web application that mitigates the challenges of labor-intensive data annotation in computer vision through AI -assistance. By integrating state-of-the-art AI models into this tool, images can be efficiently pre-annotated while proposing some labels labeling workflow. Focused on object detection, oriented bounding box estimation, and instance segmentation, VisioFirm's core lies in its pre annotation pipeline that combines pretrained detectors and zero-shot models. This framework employs low-confidence thresholding to maximize recall, followed by CLIP-based semantic verification on cropped regions. Redundancies are then mitigated through IoU-Graph Connected Components Clustering, where overlapping detections ($IoU > 0.9$) are grouped and resolved, ensuring minimal duplicates while preserving accuracy. Browser-side acceleration via WebGPU for SAM2-guided segmentation, supporting on-the-fly refinements with adaptive Ramer-Douglas-Peucker contour simplification. Flexible user options for refinement: in-app live filtering or blind trust of pre-annotations, with seamless transitions to semi-automated or manual modes for complex scenarios (unusual objects or textures).

The future directions aim to expand VisioFirm's scope and compatibility. Integration with frameworks like Detectron2 will enable advanced instance segmentation workflows. Additional CV-related annotations, such as image classification and captioning, will be incorporated to support multimodal tasks. Video support, including frame extraction and tracking-based annotation, will extend the tool to temporal data.

\bibliographystyle{plain}  
\bibliography{references}  



\end{document}